# Learning Stability Certificate for Robotics in Real-World Environments

Zhe Shen, *Member, IEEE*

*Abstract*— Stability certificates play a critical role in ensuring the safety and reliability of robotic systems. However, deriving these certificates for complex, unknown systems has traditionally required explicit knowledge of system dynamics, often making it a daunting task. This work introduces a novel framework that learns a Lyapunov function directly from trajectory data, enabling the certification of stability for autonomous systems without needing detailed system models. By parameterizing the Lyapunov candidate using a neural network and ensuring positive definiteness through Cholesky factorization, our approach automatically identifies whether the system is stable under the given trajectory. To address the challenges posed by noisy, real-world data, we allow for controlled violations of the stability condition, focusing on maintaining high confidence in the stability certification process. Our results demonstrate that this framework can provide data-driven stability guarantees, offering a robust method for certifying the safety of robotic systems in dynamic, real-world environments. This approach works without access to the internal control algorithms, making it applicable even in situations where system behavior is opaque or proprietary.

**The tool for learning the stability proof is open-sourced by this research: https://github.com/HansOersted/stability.**

## I. Introduction

The control module has always been one of the most crucial components in robotic systems. Whether it's a robotic arm [1], an autonomous driving system, or any other type of robot [2], imperfect controller logic can potentially lead to dangerous situations. Indeed, during the controller design process, we generally aim to find stability criteria [3], allowing us to theoretically prove that the designed controller is capable of stabilizing the robot. This ensures that the robot can perform tasks as intended, safely and efficiently, while avoiding potential risks due to instability.

However, deriving stability criteria for certain controllers can be mathematically challenging. For example, the stability proof of a geometric controller [4] is not straightforward because it involves highly advanced mathematics. As a result, stability criteria are often confined to a small group of experts, such as control scientists. This has led to a situation where robotic engineers may neglect the stability of controllers.

Additionally, finding analytical stability criteria for some controllers is either difficult or impossible, such as those based on neural networks or reinforcement learning [5]. We cannot use a unified formula to describe the tracking error behavior of these types of controllers. While we can ensure that, to a certain extent, the controllers will be effective and reliable with a certain probability, the designs still might lead to unintended actions—actions that, even without considering environmental noise, may cause the robot to exhibit unsafe, undefined behaviors.

In recent years, data-driven methods for exploring Lyapunov functions [6] have rapidly gained traction, disrupting traditional approaches.

By harnessing state and trajectory data, these methods eliminate the need for explicit analytical models, and in some cases, have uncovered Control Lyapunov Functions (CLFs) that certify stability for complex systems [6], [7].

Even more exciting, by tapping into recent tracking history—such as time-stamped sequences of references and system states—we can now automatically generate stability certificates.

This paradigm shift moves far beyond the limitations of model-dependent methods, opening the door to real-time verification of black-box and data-driven controllers [8], and setting the stage for a new era in controller assurance.

As of now, there has been no open-source framework capable of learning stability certificates directly from real-world, noisy data—until now.

This work [9] takes the bold step of breaking new ground by leveraging neural networks to automatically discover Input-to-State Stability (ISS) certificates, filling a critical gap in the field and opening up a new frontier for controller verification.

## II. Preliminary

In this study, we define stability through the concept of Input-to-State Stability (ISS), introduced by Sontag [10].

At the heart of our approach lies a bold demand for a positive-definite Lyapunov candidate, $V$, which must satisfy the critical inequality:

$$\dot{V} \leq \varepsilon, \qquad (1)$$

where $V(0) = 0$ and $\varepsilon$ is a positive constant waiting to be uncovered.

Building on revolutionary ideas from previous work [11], we boldly structure our Lyapunov candidate as follows:

$$V(\xi) = \xi^\top Q \xi, \qquad (2)$$

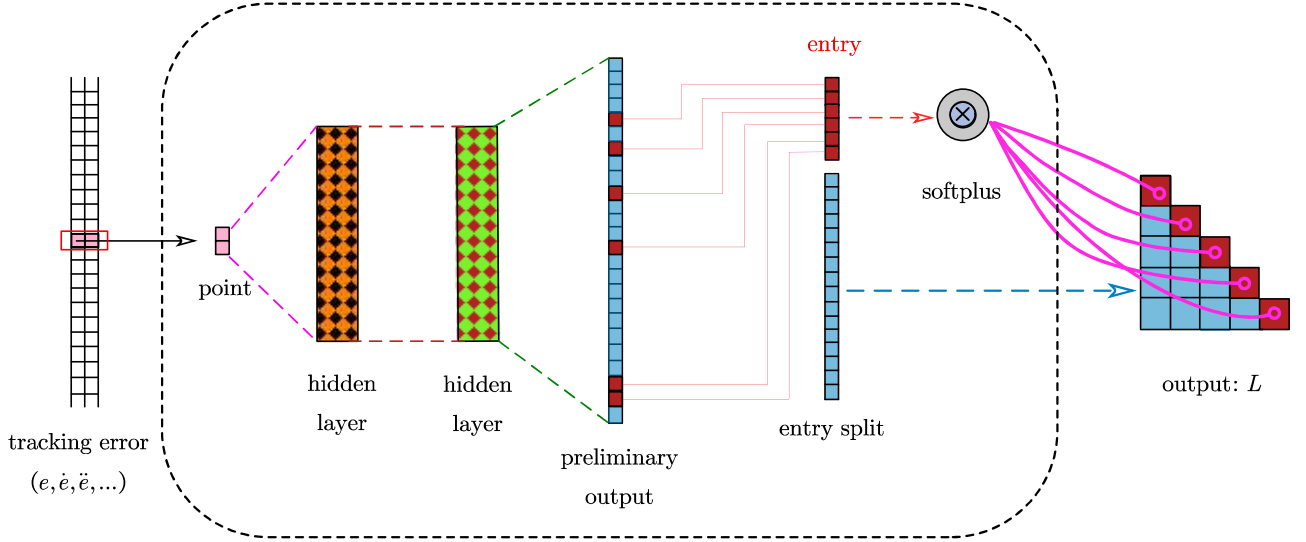

**Figure 1.** Neural network for learning the Cholesky factor. The inputs are the tracking error and its derivatives, and the output is the estimated Cholesky factor, $L$, which constitutes the parameter $Q = LL^\top$ used in the Lyapunov candidate (2).

where $\xi$ represents the state vector, and $Q$ is a symmetric positive-definite matrix. This formulation guarantees the positive-definiteness required in Equation (1) and sets the stage for unprecedented strides in stability verification.

### III. PROBLEM STATEMENT

Given the observed state trajectories, $x(t)$ and the reference trajectory, $r(t)$, which the system is intended to track.

Define the tracking error, $e(t)$, below:

$$e(t) = r(t) - x(t). \tag{3}$$

The challenge, then, is stated as: identify a Lyapunov candidate $V$, structured as in Equation (2), that satisfies the rigorous Input-to-State Stability (ISS) criterion outlined in Equation (1). This task directly impacts the stability assurance of autonomous systems.

### IV. PROPOSED FRAMEWORK

This section introduces a model-free pipeline for certifying stability directly from data. From recent, time-stamped trajectories of references and system states, we train a neural Lyapunov function and verify an ISS-style condition $\dot{V} \leq \varepsilon$ with $V(0) = 0$.

The procedure requires no explicit dynamics, tolerates measurement noise, and enables real-time verification of black-box, data-driven controllers for robotic systems.

#### A. Lyapunov Candidate Structure

To enforce positive definiteness, we re-parameterize $Q$ via a Cholesky factor:

$$Q = LL^\top, \tag{4}$$

where $L$ is a real lower triangular matrix with strictly positive diagonal entries.

Further, define the state-related term

$$\xi = \begin{bmatrix} e \\ \dot{e} \end{bmatrix}, \tag{5}$$

where $e$ is defined in Formula (3).

#### B. Loss Function

Define the loss function

$$\psi(\xi) = \max\{0, h(\xi)\}, \tag{6}$$

where

$$h(\xi) = \dot{V}(\xi) + \gamma, \tag{7}$$

where $\gamma > 0$ is a small positive constant and $\dot{V}(\xi)$ is computed according to (2), (4), and (5).

At each training epoch, the neural network is trained to minimize this loss by updating its weights and biases via gradient-based optimization.

#### C. Neural Network Architecture

The adopted architecture of the Neural Network is depicted in Fig. 1 [11], which maps the tracking error defined in (3) through several hidden layers to a vector of unconstrained outputs.

The tracking error defined in (3) is passed through hidden layers, producing a preliminary output whose elements are unconstrained, i.e., they may take either positive or negative values.

This vector is partitioned into two sets:

(i) the off-diagonal entries of a real lower-triangular matrix are taken directly;

(ii) the $n$ diagonal entries are passed through a Softplus transformation. The assembled matrix $L$ is therefore lower-triangular with strictly positive diagonal elements, guaranteeing $Q = LL^\mathsf{T} > 0$.

For further implementation details, the open-source code is available at https://github.com/HansOersted/stability.

## V. Conclusion

The proposed neural network identifies a Lyapunov function that certifies system stability under the stated ISS condition.

For further introduction, more implementation details, and more cases, we refer to our previous work [9].

## Code and Contribution

The code in this research is open source and available at: https://github.com/HansOersted/stability.

The code is released under an academic non-commercial license, with copyright © 2025 Zhe Shen and distributed through Stable Robotics Ltd., United Kingdom.

## References

[1] Y. Park and C. Sloth, "Differential-Algebraic Equation Control Barrier Function for Flexible Link Manipulator".
[2] Z. Shen and T. Tsuchiya, "Tracking Control for a Tilt-rotor with Input Constraints by Robust Gaits," in 2023 IEEE Aerospace Conference, IEEE, 2023, pp. 1–7. Accessed: Nov. 12, 2023. [Online]. Available: https://ieeexplore.ieee.org/abstract/document/10115566/
[3] A. Lavaei and D. Angeli, "Data-Driven Stability Certificate of Interconnected Homogeneous Networks via ISS Properties," IEEE Control Systems Letters, vol. 7, pp. 2395–2400, 2023, doi: 10.1109/LCSYS.2023.3285753.
[4] T. Lee, M. Leok, and N. H. McClamroch, "Geometric tracking control of a quadrotor UAV on SE(3)," in 49th IEEE Conference on Decision and Control (CDC), Atlanta, GA: IEEE, Dec. 2010, pp. 5420–5425. doi: 10.1109/CDC.2010.5717652.
[5] F. Berkenkamp, M. Turchetta, A. Schoellig, and A. Krause, "Safe Model-based Reinforcement Learning with Stability Guarantees," in Advances in Neural Information Processing Systems, Curran Associates, Inc., 2017. Accessed: Jun. 29, 2024. [Online]. Available: https://proceedings.neurips.cc/paper_files/paper/2017/hash/766ebcd59621e305170616ba3d3dac32-Abstract.html
[6] A. Lavaei, P. M. Esfahani, and M. Zamani, "Data-Driven Stability Verification of Homogeneous Nonlinear Systems with Unknown Dynamics," in 2022 IEEE 61st Conference on Decision and Control (CDC), Cancun, Mexico: IEEE, Dec. 2022, pp. 7296–7301. doi: 10.1109/CDC51059.2022.9992739.
[7] N. Boffi, S. Tu, N. Matni, J.-J. Slotine, and V. Sindhwani, "Learning Stability Certificates from Data," in Proceedings of the 2020 Conference on Robot Learning, PMLR, Oct. 2021, pp. 1341–1350. Accessed: Jan. 20, 2025. [Online]. Available: https://proceedings.mlr.press/v155/boffi21a.html
[8] A. Balkan, P. Tabuada, J. V. Deshmukh, X. Jin, and J. Kapinski, "Underminer: a framework for automatically identifying non-converging behaviors in black box system models," in Proceedings of the 13th International Conference on Embedded Software, Pittsburgh Pennsylvania: ACM, Oct. 2016, pp. 1–10. doi: 10.1145/2968478.2968487.
[9] Z. Shen, "The First Open-Source Framework for Learning Stability Certificate from Data," Sep. 23, 2025, arXiv: arXiv:2509.20392. doi: 10.48550/arXiv.2509.20392.
[10] E. D. Sontag, "Input to state stability: Basic concepts and results," in Nonlinear and Optimal Control Theory, Springer, 2008, pp. 163–220.
[11] Z. Shen, Y. Kim, and C. Sloth, "Towards Data-Driven Model-Free Safety-Critical Control," Jun. 07, 2025, arXiv: arXiv:2506.06931. doi: 10.48550/arXiv.2506.06931.